\documentclass{article} 
\usepackage{arxiv}
\usepackage{times}

\usepackage[utf8]{inputenc} 
\usepackage[T1]{fontenc}    
\usepackage{hyperref}       
\usepackage{url}            
\usepackage{booktabs}       
\usepackage{amsfonts}       
\usepackage{nicefrac}       
\usepackage{microtype}      

\usepackage{graphicx}
\usepackage{amsmath}
\usepackage{amssymb}
\usepackage{booktabs}
\usepackage[utf8]{inputenc} 
\usepackage[T1]{fontenc}    
\usepackage{url}            
\usepackage{amsfonts}       
\usepackage{nicefrac}       
\usepackage{microtype}      
\usepackage{bbding}
\usepackage{pifont}
\usepackage{bbm}
\usepackage{bm}
\usepackage{soul}
\usepackage{multirow}
\usepackage{framed}
\usepackage{amsthm}
\usepackage{makecell}
\usepackage{subfigure}
\usepackage[ruled,linesnumbered]{algorithm2e}
\usepackage{hyperref}

\usepackage{amsmath}
\usepackage{amssymb}
\usepackage{mathtools}
\usepackage{amsthm}

\usepackage[capitalize,noabbrev]{cleveref}

\theoremstyle{plain}
\newtheorem{theorem}{Theorem}

\theoremstyle{definition}

\makeatletter
\newtheoremstyle{namedtheoremstyle}%
  {\topsep}{\topsep}
  {\itshape}{}
  {\bfseries}{.}
  {0.5em}
  {\thmname{\@ifempty{#3}{#1}\@ifnotempty{#3}{#3}}}
\makeatother

\theoremstyle{namedtheoremstyle}

\usepackage{wrapfig}
\usepackage{caption}

\allowdisplaybreaks

\title{Technical Note:\\[5pt]
Defining and Quantifying AND-OR Interactions\\[3pt]
for Faithful and Concise Explanation of DNNs}

\author{\textbf{Mingjie Li}\quad \textbf{Quanshi Zhang}\thanks{Quanshi Zhang is the corresponding author. He is with the Department of Computer Science and Engineering, the John Hopcroft Center, at the Shanghai Jiao Tong University, China. \texttt{<zqs1022@sjtu.edu.cn>}}\\[3pt]
Shanghai Jiao Tong University\\
}

\begin{document}

\maketitle

\begin{abstract}
This technical note is just a short summarizaiton of partial techniques of extracting AND-OR interactions in the paper “anonymous authors, Explaining the Exact QiGan Encoded by a DNN to Play the Go Game.” 

In this technical note, we aim to explain a deep neural network (DNN) by quantifying the encoded interactions between input variables, which reflects the DNN's inference logic.
Specifically, we first rethink the definition of interactions, and then formally define \textit{faithfulness} and \textit{conciseness} for interaction-based explanation.
To this end,  we propose two kinds of interactions, \emph{i.e.}, the AND interaction and the OR interaction.
For faithfulness, we prove the uniqueness of the AND (OR) interaction in quantifying the effect of the AND (OR) relationship between input variables.
Besides, based on AND-OR interactions, we design techniques to boost the conciseness of the explanation, while not hurting the faithfulness.
In this way, the inference logic of a DNN can be faithfully and concisely explained by a set of symbolic concepts.
\end{abstract}

\section{Introduction}


This technical note is just a short summarizaiton of partial techniques of extracting AND-OR interactions in the paper “anonymous authors, Explaining the Exact QiGan Encoded by a DNN to Play the Go Game.”

Deep neural networks (DNNs) have shown promise in various tasks in recent years, but the black-box nature of a DNN makes it difficult for people to understand its internal behavior.
Many post-hoc explanation methods have been proposed either to explain the DNN visually~\citep{yosinski2015understanding,simonyan2013deep,zeiler2014visualizing}, or to estimate the attribution/importance of input variables~\citep{ribeiro2016should,lundberg2017unified,selvaraju2017grad}.

Beyond the above post-hoc explanations at the pixel-wise and unit-wise levels, in this study, we aim to explain the inference logic of a DNN by quantifying the encoded interactions between input variables.
Actually, defining and quantifying interactions is originally a typical research direction in game theory~\citep{grabisch1999axiomatic,sundararajan2020shapley,tsai2022faith}.

In recent years, our research group led by Dr. Quanshi Zhang has developed a theoretical system based on the game-theoretical interaction to solve two major challenges in explainable AI, \emph{i.e.}, 
(1) how to define and quantify implicit knowledge encoded by a DNN as explicit and countable concepts~\citep{zhang2021building,zhang2021interpreting,chengxu2021concepts,chengxu2021hypothesis,ren2023defining,ren2023can,li2023transferability,shen2023can}, and
(2) how to use concepts encoded by the DNN to explain its representation power or performance~\citep{zhang2021dropout,wangxin2021interpreting,ren2021game,wangxin2021unified,zhouhuilin2023generalization,deng2022discovering,deng2023BNN}. 
More importantly, the above theory also provides a good perspective to analyze the common mechanism shared by previous empirical findings and explanations of DNNs~\citep{deng2021unified,deng2022unify, zhangquanshi2022proving}.
The game-theoretic interactions can also help people design inherently interpretable models~\citep{chen2023harsanyinet}.

Let us first understand interactions between input variables encoded by a DNN.
Due to the non-linearity, a DNN does not independently use each individual input variables for inference like a linear model.
Instead, the DNN usually encodes interactions between different input variables for inference.
For example, given the sentence ``\textit{sit down and take it easy},'' the DNN does not directly use individual words for inference.
Instead, different words may collaborate with each other to form meaningful phrases used in the DNN's inference.
As an example, the interaction between the three words {\small$\{\textit{take, it, easy}\}$} forms a phrase meaning ``calm down'', and makes a specific numerical contribution to the model output, which is termed the \textbf{\textit{interaction effect}}.

In this study, we quantify two types of interactions between input variables based on the collaboration relationship between input variables.
Then, we explain the inference logic of the DNN by decomposing the DNN's output into the interaction effects.

\textit{(1) The AND interaction.}
Again, let us consider the above sentence ``\textit{sit down and take it easy}'' as an example.
In this sentence, only when the three words in {\small$S=\{\textit{take, it, easy}\}$} co-appear, they will contribute the meaning of ``\textit{calm down}'' to the DNN's output.
Therefore, words ``\textit{take}'', ``\textit{it}'', and ``\textit{easy}'' form an \textbf{\textit{AND relationship}}, meaning that only their co-appearance contributes an interaction effect {\small$\mathcal{I}^{\textit{AND}}(S)$} to the DNN's output.
On the other hand, the absence (masking) of any words in {\small$\{\textit{take, it, easy}\}$} will remove the effect {\small$\mathcal{I}^{\textit{AND}}(S)$} from the DNN's output.
Such an interaction between input variables in {\small$S$} is termed an \textbf{\textit{AND interaction}}.

\textit{(2) The OR interaction.}
Let us consider another sentence ``\textit{I think this movie is boring and disappointing}''.
In this sentence, if any word in {\small$S=\{\textit{boring, disappointing}\}$} appears, it will contribute a negative effect to the overall sentiment.
Therefore, words ``\textit{boring}'' and ``\textit{disappointing}'' form an \textbf{\textit{OR relationship}}, meaning that the appearance of any word in {\small$S$} will contribute an interaction effect {\small$\mathcal{I}^{\textit{OR}}(S)$} to the DNN's output.
In this case, only when all words in {\small$\{\textit{boring, disappointing}\}$} are masked, the effect {\small$\mathcal{I}^{\textit{OR}}(S)$} can be removed from the DNN's output.
Such an interaction between input variables in {\small$S$} is termed an \textbf{\textit{OR interaction}}.

However, using the above AND-OR interactions to explain deep models poses new distinctive issues, which are beyond previous issues on interactions.
Many classical interactions~\cite{grabisch1999axiomatic,sundararajan2020shapley} were defined in game theory, and they were mainly proposed to satisfy some desirable game-theoretic axioms.
\textit{In comparison, besides such game-theoretic axioms, faithfulness and conciseness are two distinctive requirements for explanation of DNNs.}

$\bullet$~\textbf{Faithfulness.}
Given an input sample with {\small$n$} variables, there are {\small$2^n$} different ways to randomly mask input variables.
Given any one of all the {\small$2^n$} masked input samples, if the DNN's output can always be decomposed into all triggered interaction effects, then we can consider the set of interaction effects as a faithful explanation for the inference logic of the DNN.

$\bullet$~\textbf{Conciseness.}
Besides faithfulness, the interaction-based explanation needs to be concise enough for people to understand.
However, theoretically, given a DNN with $n$ input variables, we may extract at most {\small $2^n$} possible AND interactions and {\small $2^n$} possible OR interactions, which is too complex for explanation.
To this end, we propose a method to boost the sparsity of interaction effects in explanation while not hurting the faithfulness, so that the inference score of a DNN can be concisely explained by a small set of AND-OR interactions.

$\bullet$~\textbf{Concept discovery.}
Since we can use a small set of AND-OR interactions to faithfully explain the inference logic of a DNN, each interaction can be considered as a specific symbolic concept encoded by the deep model. 

In fact, this paper is just a technical note for AND-OR interactions.
In papers mentioned above, we have formally introduced details of such interaction-based explanation, provided mathematical proofs of its theoretical properties, and verified the effectiveness of this method in real applications.

\section{Algorithm}
\label{sec:algo}

\subsection{The AND interaction: decomposing AND relationship between input variables}
\label{subsec:and-interaction}

Given a pre-trained deep model {\small$v(\cdot)$} and an input sample {\small$x$}, let us introduce how to represent the model output {\small$v(x)\in\mathbb{R}$} as a set of AND interactions.
Let us take a model for sentiment classification and the input sentence ``\textit{sit down and take it easy}'' for example.
Each word in this sentence can be taken as an input variable to the model.
Let {\small$N=\{\textit{sit, down, and, take, it, easy}\}$} denote the set of all input variables.
Input variables in {\small$N$} do not contribute to the model output {\small$v(x)$} individually.
Instead, they interact with each other to form meaningful patterns for inference.
For example, in this sentence, the co-appearance of the three words {\small$S=\{\textit{take,it,easy}\}$} would contribute a significant additional effect that pushes the model output towards the positive meaning ``calm down'', compared with a sentence with any words being masked, \emph{e.g.,} {\small$\{\textit{take}, \textit{it}, \text{\_\_}\}$}, {\small$\{\textit{take}, \text{\_\_}, \textit{easy}\}$}, {\small$\{\text{\_\_}, \textit{it}, \textit{easy}\}$}.
Therefore, we can consider words in {\small$S$} form an AND relationship, \emph{i.e.} only their co-appearance will contribute an additional effect {\small$\mathcal{I}^{\textit{AND}}(S)$} to the model output.

Mathematically, the interaction effect of the AND relationship among a set of input variables {\small$S\subseteq N$}, given as {\small$\mathcal{I}^{\textit{AND}}(S)$}, quantifies the marginal numerical effect of the co-appearance of variables in {\small$S$} on the model output {\small$v(x_S)$}.
Specifically, {\small$v(x_S),S\subseteq N$} denotes the output score of the model when only variables in {\small$S$} are given, and all variables in {\small$N\backslash S$} are masked using their baseline values~\citep{dabkowski2017real,ancona2019explaining} in {\small$\boldsymbol{r}=[r_1,r_2,\ldots,r_n]$}, as follows.

\begin{equation}
     (x_S)_i=
     \left\{
     \begin{array}{ll}
        \!\! x_i,  & i\in S \\
        \!\! r_i,  & i\in N\!\setminus\! S
     \end{array}
     \right.
\label{eq:def-of-vs}
\end{equation}

For simplicity, we denote {\small$v(x_S)$} as {\small$v(S)$}. 
In particular, {\small$v(N)$} denote the model output given the original sample {\small$x$} as the input.
{\small$v(\emptyset)$} denote the model output when all input variables are masked by baseline values.

\textbf{Quantifying AND interactions that satisfy faithfulness.}
Then, we aim to decompose the output of a DNN into AND interactions.
In this paper, we consider that a metric for AND interactions is faithful if and only if we can always use the triggered effects {\small$\mathcal{I}^{\textit{AND}}(S)$} of AND interactions to reconstruct the model output {\small$v(T)$} given any subset of input variables {\small$T\subseteq N$}, as follows.

\begin{equation}
    \forall T\subseteq N,\ \  v(T)={\sum}_{S\subseteq T} \mathcal{I}^{\textit{AND}}(S)
    \label{eq:and-faith}
\end{equation}

To this end, \citet{ren2023defining} have proved that the Harsanyi dividend~\citep{harsanyi1963simplified} is the unique metric for AND interactions that fully satisfies the faithfulness requirement in Eq. (\ref{eq:and-faith}), \emph{i.e.} {\small$\forall T\subseteq N, v(T)={\sum}_{S\subseteq T} \mathcal{I}^{\textit{AND}}(S)$}.
Therefore, we use Harsanyi dividends to quantify effects of AND interactions encoded by a DNN, as follows.

\begin{equation}
    \mathcal{I}^{\textit{AND}}(S)={\sum}_{L\subseteq S}(-1)^{|S|-|L|}v(S)
\label{eq:def-and-interaction}
\end{equation}

\textbf{Axioms and theorems.}
Besides, \citet{ren2023defining} have also proved that the interaction effect based on the Harsanyi dividend satisfies the following desirable axioms, which enhance the trustworthiness of this metric.

(1) \textit{Efficiency axiom} (proved by \citet{harsanyi1963simplified}). The output score of a model can be decomposed into interaction utilities of different patterns, \emph{i.e.} {\small $v(N)=\sum_{S\subseteq N}\mathcal{I}^{\textit{AND}}(S)$}.\\[2pt]
(2) \textit{Linearity axiom}. If we merge output scores of two models $w$ and $v$ as the output of model $u$, \emph{i.e.} {\small $\forall S\subseteq N,~ u(S)=w(S)+v(S)$}, then their interaction utilities {\small $\mathcal{I}^{\textit{AND}}_v(S)$} and {\small $\mathcal{I}^{\textit{AND}}_w(S)$} can also be merged as {\small $\forall S\subseteq N, \mathcal{I}^{\textit{AND}}_u(S)=\mathcal{I}^{\textit{AND}}_v(S)+\mathcal{I}^{\textit{AND}}_w(S)$}.\\[2pt]
(3) \textit{Dummy axiom}. If a variable {\small $i\in N$} is a dummy variable, \emph{i.e.} 
{\small $\forall S\subseteq N\backslash\{i\}, v(S\cup\{i\})=v(S)+v(\{i\})$}, then it has no interaction with other variables, {\small $\forall S\subseteq N\backslash\{i\}, \mathcal{I}^{\textit{AND}}(S\cup\{i\})=0$}.\\[2pt]
(4) \textit{Symmetry axiom}. If input variables {\small $i,j\in N$} cooperate with other variables in the same way, {\small $\forall S\subseteq N\backslash\{i,j\}, v(S\cup\{i\})=v(S\cup\{j\})$}, then they have same interaction utilities with other variables, {\small $\forall S\subseteq N\backslash\{i,j\}, \mathcal{I}^{\textit{AND}}(S\cup\{i\})=\mathcal{I}^{\textit{AND}}(S\cup\{j\})$}.\\[2pt]
(5) \textit{Anonymity axiom}. For any permutations $\pi$ on {\small $N$}, we have {\small $\forall S\subseteq N, \mathcal{I}^{\textit{AND}}_{v}(S)=\mathcal{I}^{\textit{AND}}_{\pi v}(\pi S)$}, where {\small $\pi S\triangleq\{\pi(i)|i\in S\}$}, and the new model {\small $\pi v$} is defined by {\small $(\pi v)(\pi S)=v(S)$}.
This indicates that interaction utilities are not changed by permutation.\\[2pt]
(6) \textit{Recursive axiom}. The interaction utilities can be computed recursively.
For {\small $i\in N$} and {\small $S\subseteq N\backslash\{i\}$}, the interaction utility of the pattern {\small $S\cup\{i\}$} is equal to the interaction utility of {\small $S$} with the presence of $i$ minus the interaction utility of $S$ with the absence of $i$, \emph{i.e.} {\small $\forall S\subseteq N\setminus\{i\}, \mathcal{I}^{\textit{AND}}(S\cup \{i\})=\mathcal{I}^{\textit{AND}}(S|i\text{ is always present})-\mathcal{I}^{\textit{AND}}(S)$}. {\small $\mathcal{I}^{\textit{AND}}(S|i\text{ is always present})$} denotes the interaction utility when the variable $i$ is always present as a constant context, \emph{i.e.} {\small$
\mathcal{I}^{\textit{AND}}(S|i\text{ is always present})=\sum_{L\subseteq S} (-1)^{|S|-|L|}\cdot v(L\cup\{i\})$}.\\[2pt]
(7) \textit{Interaction distribution axiom}. This axiom characterizes how interactions are distributed for ``interaction functions''~\citep{sundararajan2020shapley}.
An interaction function {\small $v_T$} parameterized by a context {\small $T$} satisfies {\small $\forall S\subseteq N$}, if {\small $T\subseteq S$}, {\small$v_T(S)=c$}; otherwise, {\small $v_T(S)=0$}. Then, we have {\small $\mathcal{I}^{\textit{AND}}(T)=c$}, and {\small $\forall S\neq T$}, {\small $\mathcal{I}^{\textit{AND}}(S)=0$}.

Furthermore, \citet{ren2023defining} have also proven that the interaction effects {\small $\mathcal{I}^{\textit{AND}}(S)$} can explain the elementary mechanism of existing game-theoretic attributions/interactions.

\begin{theorem}[Connection to the Shapley value, proved by \citet{harsanyi1963simplified}]
Let {\small $\phi(i)$} denote the Shapley value~\citep{shapley1953value} of an input variable $i$. Then, its Shapley value can be represented as the weighted sum of interaction utilities, \emph{i.e.} {\small $\phi(i)=\sum_{S\subseteq N\backslash\{i\}}\frac{1}{|S|+1} \mathcal{I}^{\textit{AND}}(S\cup\{i\})$}.
In other words, the utility of an interaction pattern with $m$ variables should be equally assigned to the $m$ variables in the computation of Shapley values.
\end{theorem}

\begin{theorem}[Connection to the Shapley interaction index]
Given a subset of input variables {\small $T\subseteq N$}, {\small $\mathcal{I}^{\textrm{Shapley}}(T)=\sum_{S\subseteq N\backslash T}\frac{|S|!(|N|-|S|-|T|)!}{(|N|-|T|+1)!}\Delta v_T(S)$} denotes the Shapley interaction index~\citep{grabisch1999axiomatic} of {\small $T$}. We have proven that the Shapley interaction index can be represented as the weighted sum of  utilities of interaction patterns, {\small $\mathcal{I}^{\textrm{Shapley}}(T)=\sum_{S\subseteq N\backslash T}\frac{1}{|S|+1}\mathcal{I}^{\textit{AND}}(S\cup T)$}.
This metric treats the  coalition of variables in {\small $T$} as a single variable, and thus uniformly allocates the attribution {\small $\mathcal{I}^{\textit{AND}}(S\cup T)$} to all variables including {\small $T$}.
\end{theorem}

\begin{theorem}[Connection to the Shapley Taylor interaction index]
Given a subset of input variables {\small $T\subseteq N$}, let {\small $\mathcal{I}^{\textrm{Shapley-Taylor}}(T)$} denote the Shapley Taylor interaction index~\citep{sundararajan2020shapley} of order $k$ for {\small $T$}.
We have proven that the Shapley taylor interaction index can be represented as the weighted sum of interaction utilities, \emph{i.e.} {\small $\mathcal{I}^{\textrm{Shapley-Taylor}}(T)=\mathcal{I}^{\textit{AND}}(T)$} if {\small $|T|<k$}; {\small $\mathcal{I}^{\textrm{Shapley-Taylor}}(T)=\sum_{S\subseteq N\backslash T}\binom{|S|+k}{k}^{-1}\mathcal{I}^{\textit{AND}}(S\cup T)$} if {\small $|T|=k$}; and {\small $\mathcal{I}^{\textrm{Shapley-Taylor}}(T)=0$ if $|T|>k$}.
\end{theorem}

\subsection{The OR interaction: extending to OR relationship between input variables}
\label{subsec:or-interaction}

In this section, besides AND interactions, we find that the model output can also be equivalently represented as OR interactions. 
Let us consider another input sentence ``\textit{I think this movie is boring and disappointing}'' for sentiment classification.
In this sentence, if any word in {\small$S=\{\textit{boring, disappointing}\}$} appears, it would contribute a significant effect that pushes the model output towards the negative sentiment.
Therefore, we can consider words in {\small$S$} form an OR relationship, \emph{i.e.}, the appearance of any word will contribute an additional effect {\small$\mathcal{I}^{\textit{OR}}(S)$} to the model output.

\textbf{Quantifying OR interactions that satisfy faithfulness.}
Just like the quantification of AND interactions in Section~\ref{subsec:and-interaction}, we consider that a metric for OR interactions is faithful if we can always use the triggered effects {\small$\mathcal{I}^{\textit{OR}}(S)$} to reconstruct the model output {\small$v(T)$} given any subset of input variables {\small$T\subseteq N$}.
The following equation shows the faithfulness requirement for OR interactions.

\begin{equation}
    \forall T\subseteq N,\ \  v(T)=\mathcal{I}^{\textit{OR}}(\emptyset)+{\sum}_{S\cap T\neq\emptyset} \mathcal{I}^{\textit{OR}}(S)
    \label{eq:or-faith}
\end{equation}

To this end, in this study, we prove that the following metric is the unique metric for OR interactions {\small$\mathcal{I}^{\textit{OR}}(S)$}, which fully satisfies the faithfulness requirement for OR interactions in Eq. (\ref{eq:or-faith}).

\begin{equation}
    \mathcal{I}^{\textit{OR}}(S)=\left\{
    \begin{array}{ll}
        v(\emptyset),\ & S=\emptyset \\[5pt]
        -\sum_{L\subseteq S}(-1)^{|S|-|L|}v(N\backslash L),\ & S\neq\emptyset
    \end{array}
    \right.
\label{eq:def-or-interaction}
\end{equation}

\subsection{Faithfully and concisely explaining the DNN based on AND-OR interactions}
\label{subsec:and-or-interaction}

In fact, we believe that the model output is a mixture model of both AND interactions and OR interactions.
In this section, we aim to disentangle the model output into both AND interactions and OR interactions, and seek the best combination of AND interactions and OR interactions that leads to the most concise explanation.
In other words, we aim to faithfully decompose the model output into a small set of interactions, which consists of both AND interactions and OR interactions.

\textbf{Faithfulness.}
Just like in Section~\ref{subsec:and-interaction} and Section~\ref{subsec:or-interaction}, the faithfulness of such interaction-based explanation is guaranteed by the following requirement, \emph{i.e.} the model output {\small$v(T)$} given any subset of input variables {\small$T\subseteq N$} can always be reconstructed by the sum of all triggered effects of AND-OR interactions, as follows.

\begin{equation}
    \forall T\subseteq N,\ \ \ v(T)=\ 
    \underbrace{{\sum}_{S\subseteq T} \hat{\mathcal{I}}^{\textit{AND}}(S)}_{\substack{\text{the sum of activated}\\[2pt] \text{AND interaction effects}}}
    \ \ +\ \ 
    \underbrace{\hat{\mathcal{I}}^{\textit{OR}}(\emptyset)+{\sum}_{S\cap T\neq\emptyset} \hat{\mathcal{I}}^{\textit{OR}}(S)}_{\substack{\text{the sum of activated}\\[2pt] \text{OR interaction effects}}},
\label{eq:and-or-faith}
\end{equation}

where {\small$\hat{\mathcal{I}}^{\textit{AND}}(S)$} represents the effect of the AND interaction in {\small$S$}, and {\small$\hat{\mathcal{I}}^{\textit{OR}}(S)$} represents the effect of the OR interaction in {\small$S$}.

\textbf{Conciseness.}
In order to boost the conciseness of the above explanation, we optimize the AND interactions {\small$\{\hat{\mathcal{I}}^{\textit{AND}}(S)\}_{S\subseteq N}$} and the OR interactions {\small$\{\hat{\mathcal{I}}^{\textit{OR}}(S)\}_{S\subseteq N}$} using a Lasso loss, while forcing the faithfulness constraint in Eq. (\ref{eq:and-or-faith}) to hold.

\begin{equation}
\begin{aligned}
    \min_{\substack{\{\hat{\mathcal{I}}^{\textit{AND}}(S)\}_{S\subseteq N}\\\{\hat{\mathcal{I}}^{\textit{OR}}(S)\}_{S\subseteq N}}}\ \  & \sum_{S\subseteq N}\ \big|\hat{\mathcal{I}}^{\textit{AND}}(S)\big| + \sum_{S\subseteq N}\ \big|\hat{\mathcal{I}}^{\textit{OR}}(S)\big|  \\ 
    \mathrm{s.t.}\qquad & \ \ \text{Eq. (\ref{eq:and-or-faith}) holds}
\end{aligned}
\label{eq:conciseness-vanilla}
\end{equation}

The optimization problem in Eq. (\ref{eq:conciseness-vanilla}) is equivalent to explaining part of the model output {\small$v^{\textit{AND}}(\cdot)$} based on AND interactions in Eq. (\ref{eq:def-and-interaction}), and explaining the rest part of the model output {\small$v^{\textit{OR}}(\cdot)=v(\cdot)-v^{\textit{AND}}(\cdot)$} based on OR interactions in Eq. (\ref{eq:def-or-interaction}).
Therefore, the optimization problem in Eq. (\ref{eq:conciseness-vanilla}) is actually learning the best decomposition of the model output, \emph{i.e.} {\small$\forall T\subseteq N,\ v(S)=v^{\textit{AND}}(S)+v^{\textit{OR}}(S)$}.
In this way, the optimization of Eq. (\ref{eq:conciseness-vanilla}) is equivalent to the following optimization problem.

\begin{equation}
\begin{aligned}
    \min_{\substack{\{\hat{\mathcal{I}}^{\textit{AND}}(S)\}_{S\subseteq N}\\\{\hat{\mathcal{I}}^{\textit{OR}}(S)\}_{S\subseteq N}}}\ \  & \sum_{S\subseteq N}\ \big|\hat{\mathcal{I}}^{\textit{AND}}(S)\big| + \sum_{S\subseteq N}\ \big|\hat{\mathcal{I}}^{\textit{OR}}(S)\big|  \\ 
    \mathrm{s.t.}\qquad 
    & \ \ \hat{\mathcal{I}}^{\textit{AND}}(S)={\sum}_{L\subseteq S}(-1)^{|S|-|L|}v^{\textit{AND}}(S) \\[3pt]
    & \ \ 
    \hat{\mathcal{I}}^{\textit{OR}}(S)=\left\{
    \begin{array}{ll}
        v^{\textit{OR}}(\emptyset),\ & S=\emptyset \\[2pt]
        -\sum_{L\subseteq S}(-1)^{|S|-|L|}v^{\textit{OR}}(N\backslash L),\ & S\neq\emptyset
    \end{array}
    \right.\\[3pt]
    & \ \ \forall S\subseteq N,\ v(S)=v^{\textit{AND}}(S)+v^{\textit{OR}}(S)
\end{aligned}
\label{eq:conciseness-decompose}
\end{equation}

\subsection{Implementation details}
\label{subsec:and-or-detail}

In this section, we provide implementation details for optimizing Eq. (\ref{eq:conciseness-decompose}), and propose another technique to further boost the conciseness of explanation.
In Eq. (\ref{eq:conciseness-decompose}), we perform a partition on the DNN's output, \emph{i.e.} {\small$\forall S\subseteq N,\ v(S)=v^{\textit{AND}}(S)+v^{\textit{OR}}(S)$}.
Let us consider a possible scheme for such a partition, \emph{i.e.} {\small$v^{\textit{AND}}(S)=\frac{1}{2}v(S)+p(S)$} and {\small$v^{\textit{OR}}(S)=\frac{1}{2}v(S)-p(S)$}, where {\small$p(S)\in\mathbb{R}$} determines how to partition the original model output {\small$v(S)$} into {\small$v^{\textit{AND}}(S)$} and {\small$v^{\textit{OR}}(S)$}.
Let {\small$\bm{v}=[v(S_1),v(S_2),...,v(S_{2^n})]\in\mathbb{R}^{2^n}$} denote the DNN's output under all {\small$2^n$} different masked samples.
Correspondingly, let {\small$\bm{p}=[p(S_1),p(S_2),...,p(S_{2^n})]\in\mathbb{R}^{2^n}$}, {\small$\bm{\mathcal{I}}^{\textit{AND}}=[\mathcal{I}^{\textit{AND}}(S_1),\mathcal{I}^{\textit{AND}}(S_2),...,\mathcal{I}^{\textit{AND}}(S_{2^n})]\in\mathbb{R}^{2^n}$}, and {\small$\bm{\mathcal{I}}^{\textit{OR}}=[\mathcal{I}^{\textit{OR}}(S_1),\mathcal{I}^{\textit{OR}}(S_2),...,\mathcal{I}^{\textit{OR}}(S_{2^n})]\in\mathbb{R}^{2^n}$}.
Then, the optimization in Eq. (\ref{eq:conciseness-decompose}) can be equivalently transformed into the following optimization problem.

\begin{equation}
\begin{aligned}
    \min_{\bm{p}\in\mathbb{R}^{2^n}}\ &\ 
    \big\Vert\hat{\bm{\mathcal{I}}}^{\textit{AND}}\big\Vert_1 + \big\Vert\hat{\bm{\mathcal{I}}}^{\textit{OR}}\big\Vert_1 \\
    \mathrm{s.t.}\ \ \ &\ \ \hat{\bm{\mathcal{I}}}^{\textit{AND}} =\  \bm{T}^{\textit{AND}}\left[\frac{1}{2}\bm{v}+\bm{p}\right] \\
    & \ \ \ \hat{\bm{\mathcal{I}}}^{\textit{OR}}\ =\ \  \bm{T}^{\textit{OR}}\ \left[\frac{1}{2}\bm{v}-\bm{p}\right],
\end{aligned}
\label{eq:conciseness-p}
\end{equation}

where {\small$\bm{T}^{\textit{AND}}\in\{0,+1,-1\}^{2^n\times 2^n}$} is a matrix that maps {\small$\bm{v}=[v(S_1),v(S_2),...,v(S_{2^n})]$} to {\small$\bm{\mathcal{I}}^{\textit{AND}}=[\mathcal{I}^{\textit{AND}}(S_1),\mathcal{I}^{\textit{AND}}(S_2),...,\mathcal{I}^{\textit{AND}}(S_{2^n})]$}, according to Eq. (\ref{eq:def-and-interaction}).
Similarly, {\small$\bm{T}^{\textit{OR}}\in\{0,+1,-1\}^{2^n\times 2^n}$} is a matrix that maps {\small$\bm{v}=[v(S_1),v(S_2),...,v(S_{2^n})]$} to {\small$\bm{\mathcal{I}}^{\textit{OR}}=[\mathcal{I}^{\textit{OR}}(S_1),\mathcal{I}^{\textit{OR}}(S_2),...,\mathcal{I}^{\textit{OR}}(S_{2^n})]$}, according to Eq. (\ref{eq:def-or-interaction}).

\textbf{Further boosting conciseness.}
Actually, there is a trade-off between the faithfulness and the conciseness of the explanation.
To this end, if we allow a small approximation error {\small$\epsilon_i\approx 0$} when approximating the DNN's output {\small$v(S_i)$} using the set of triggered interaction effects in Eq. (\ref{eq:and-or-faith}), the conciseness of the explanation can be further boosted.
Let {\small$\bm{\epsilon}=[\epsilon_1,\epsilon_2,...,\epsilon_{2^n}]$}.
Then, we can obtain AND-OR interactions {\small$\bm{\mathcal{I}}^{\textit{AND}}=[\mathcal{I}^{\textit{AND}}(S_1),\mathcal{I}^{\textit{AND}}(S_2),...,\mathcal{I}^{\textit{AND}}(S_{2^n})]$} and {\small$\bm{\mathcal{I}}^{\textit{OR}}=[\mathcal{I}^{\textit{OR}}(S_1),\mathcal{I}^{\textit{OR}}(S_2),...,\mathcal{I}^{\textit{OR}}(S_{2^n})]$} based on the following optimization problem.

\begin{equation}
\begin{aligned}
    \min_{\bm{p},\bm{\epsilon}\in\mathbb{R}^{2^n}}\ &\ 
    \big\Vert\hat{\bm{\mathcal{I}}}^{\textit{AND}}\big\Vert_1 + \big\Vert\hat{\bm{\mathcal{I}}}^{\textit{OR}}\big\Vert_1 \\
    \mathrm{s.t.}\ \ \ &\ \ \hat{\bm{\mathcal{I}}}^{\textit{AND}} =\  \bm{T}^{\textit{AND}}\left[\frac{1}{2}(\bm{v}+\bm{\epsilon})+\bm{p}\right] \\
    & \ \ \ \hat{\bm{\mathcal{I}}}^{\textit{OR}}\ =\ \  \bm{T}^{\textit{OR}}\ \left[\frac{1}{2}(\bm{v}+\bm{\epsilon})-\bm{p}\right] \\
    & \ \ \ \forall i,\ |\epsilon_i| < \tau_i\ \ ,
\end{aligned}
\label{eq:conciseness-pq}
\end{equation}

where {\small$\tau_i>0$} is a small scalar representing the upper bound for the approximation error {\small$\epsilon_i$}.
In practice, the optimization parameters {\small$\bm{p}$} and {\small$\bm{\epsilon}$} are usually initialized as zero vectors.
The upper bound for the approximation error can be set as {\small$0.05\cdot|v(N)-v(\emptyset)|$}.

\textbf{Extracting salient symbolic concepts.}
The obtained AND-OR interactions are usually sparse. \emph{I.e.} most interactions have little influence on the output with negligible values {\small$|\mathcal{I}^{\textit{AND}}(S)|\approx 0$} (or {\small$|\mathcal{I}^{\textit{OR}}(S)|\approx 0$}), and they are termed \textbf{\textit{noisy concepts}}.
Only a few interactions have considerable effects {\small$\mathcal{I}^{\textit{AND}}(S)$} (or {\small$\mathcal{I}^{\textit{OR}}(S)$}), and they are termed \textbf{\textit{salient concepts}}.
In this way, the DNN's output on a specific sample is explained by a small set of salient concepts, which reflects the inference logic of the DNN on this input sample.

\section{Conclusion}

In this technical note, we summarize the technical details of AND-OR interactions for the explanation of DNNs.
Specifically, we formally define \textit{faithfulness} and \textit{conciseness} for interaction-based explanation.
To this end, we propose two kinds of interactions, \emph{i.e.}, the AND interaction and the OR interaction, to quantify two types of collaboration relationship between input variables.
Based on the AND-OR interactions, we also design techniques to boost the conciseness of the explanation, while preserving the faithfulness.
In this way, the inference logic of a DNN can be faithfully and concisely explained by a set of symbolic concepts.

\bibliographystyle{plainnat}
\bibliography{ref}

\begin{thebibliography}{32}
\providecommand{\natexlab}[1]{#1}
\providecommand{\url}[1]{\texttt{#1}}
\expandafter\ifx\csname urlstyle\endcsname\relax
  \providecommand{\doi}[1]{doi: #1}\else
  \providecommand{\doi}{doi: \begingroup \urlstyle{rm}\Url}\fi

\bibitem[Ancona et~al.(2019)Ancona, Oztireli, and Gross]{ancona2019explaining}
Marco Ancona, Cengiz Oztireli, and Markus Gross.
\newblock Explaining deep neural networks with a polynomial time algorithm for shapley value approximation.
\newblock In \emph{International Conference on Machine Learning}, pages 272--281. PMLR, 2019.

\bibitem[Chen et~al.(2023)Chen, Lou, Zhang, Huang, and Zhang]{chen2023harsanyinet}
Lu~Chen, Siyu Lou, Keyan Zhang, Jin Huang, and Quanshi Zhang.
\newblock Harsanyinet: Computing accurate shapley values in a single forward propagation.
\newblock \emph{arXiv preprint arXiv:2304.01811}, 2023.

\bibitem[Cheng et~al.(2021{\natexlab{a}})Cheng, Chu, Zheng, Ren, and Zhang]{chengxu2021concepts}
Xu~Cheng, Chuntung Chu, Yi~Zheng, Jie Ren, and Quanshi Zhang.
\newblock A game-theoretic taxonomy of visual concepts in dnns.
\newblock \emph{arXiv preprint arXiv:2106.10938}, 2021{\natexlab{a}}.

\bibitem[Cheng et~al.(2021{\natexlab{b}})Cheng, Wang, Xue, Liang, and Zhang]{chengxu2021hypothesis}
Xu~Cheng, Xin Wang, Haotian Xue, Zhengyang Liang, and Quanshi Zhang.
\newblock A hypothesis for the aesthetic appreciation in neural networks.
\newblock \emph{arXiv preprint arXiv::2108.02646}, 2021{\natexlab{b}}.

\bibitem[Dabkowski and Gal(2017)]{dabkowski2017real}
Piotr Dabkowski and Yarin Gal.
\newblock Real time image saliency for black box classifiers.
\newblock \emph{arXiv preprint arXiv:1705.07857}, 2017.

\bibitem[Deng et~al.(2021)Deng, Zou, Du, Chen, Feng, and Hu]{deng2021unified}
Huiqi Deng, Na~Zou, Mengnan Du, Weifu Chen, Guocan Feng, and Xia Hu.
\newblock A unified taylor framework for revisiting attribution methods.
\newblock In \emph{Proceedings of the AAAI Conference on Artificial Intelligence}, pages 11462--11469, 2021.

\bibitem[Deng et~al.(2022{\natexlab{a}})Deng, Ren, Zhang, and Zhang]{deng2022discovering}
Huiqi Deng, Qihan Ren, Hao Zhang, and Quanshi Zhang.
\newblock Discovering and explaining the representation bottleneck of dnns.
\newblock In \emph{International Conference on Learning Representations}, 2022{\natexlab{a}}.

\bibitem[Deng et~al.(2022{\natexlab{b}})Deng, Zou, Du, Chen, Feng, Yang, Li, and Zhang]{deng2022unify}
Huiqi Deng, Na~Zou, Mengnan Du, Weifu Chen, Guocan Feng, Ziwei Yang, Zheyang Li, and Quanshi Zhang.
\newblock Understanding and unifying fourteen attribution methods with taylor interactions.
\newblock \emph{arXiv preprint}, 2022{\natexlab{b}}.

\bibitem[Grabisch and Roubens(1999)]{grabisch1999axiomatic}
Michel Grabisch and Marc Roubens.
\newblock An axiomatic approach to the concept of interaction among players in cooperative games.
\newblock \emph{International Journal of game theory}, 28\penalty0 (4):\penalty0 547--565, 1999.

\bibitem[Harsanyi(1963)]{harsanyi1963simplified}
John~C Harsanyi.
\newblock A simplified bargaining model for the n-person cooperative game.
\newblock \emph{International Economic Review}, 4\penalty0 (2):\penalty0 194--220, 1963.

\bibitem[Li and Zhang(2023)]{li2023transferability}
Mingjie Li and Quanshi Zhang.
\newblock Does a neural network really encode symbolic concept?
\newblock \emph{arXiv preprint arXiv:2302.13080}, 2023.

\bibitem[Lundberg and Lee(2017)]{lundberg2017unified}
Scott~M Lundberg and Su-In Lee.
\newblock A unified approach to interpreting model predictions.
\newblock In \emph{Proceedings of the 31st international conference on neural information processing systems}, pages 4768--4777, 2017.

\bibitem[Ren et~al.(2021)Ren, Zhang, Wang, Chen, Zhou, Chen, Cheng, Wang, Zhou, Shi, and Zhang]{ren2021game}
Jie Ren, Die Zhang, Yisen Wang, Lu~Chen, Zhanpeng Zhou, Yiting Chen, Xu~Cheng, Xin Wang, Meng Zhou, Jie Shi, and Quanshi Zhang.
\newblock Towards a unified game-theoretic view of adversarial perturbations and robustness.
\newblock In M.~Ranzato, A.~Beygelzimer, Y.~Dauphin, P.S. Liang, and J.~Wortman Vaughan, editors, \emph{Advances in Neural Information Processing Systems}, volume~34, pages 3797--3810. Curran Associates, Inc., 2021.

\bibitem[Ren et~al.(2023{\natexlab{a}})Ren, Li, Chen, Deng, and Zhang]{ren2023defining}
Jie Ren, Mingjie Li, Qirui Chen, Huiqi Deng, and Quanshi Zhang.
\newblock Defining and quantifying the emergence of sparse concepts in dnns.
\newblock In \emph{IEEE Conference on Computer Vision and Pattern Recognition}, 2023{\natexlab{a}}.

\bibitem[Ren et~al.(2023{\natexlab{b}})Ren, Zhou, Chen, and Zhang]{ren2023can}
Jie Ren, Zhanpeng Zhou, Qirui Chen, and Quanshi Zhang.
\newblock Can we faithfully represent absence states to compute shapley values on a dnn?
\newblock In \emph{The Eleventh International Conference on Learning Representations}, 2023{\natexlab{b}}.

\bibitem[Ren et~al.(2023{\natexlab{c}})Ren, Deng, Chen, Lou, and Zhang]{deng2023BNN}
Qihan Ren, Huiqi Deng, Yunuo Chen, Siyu Lou, and Quanshi Zhang.
\newblock Bayesian neural networks tend to ignore complex and sensitive concepts.
\newblock \emph{arXiv preprint arXiv:2302.13095}, 2023{\natexlab{c}}.

\bibitem[Ribeiro et~al.(2016)Ribeiro, Singh, and Guestrin]{ribeiro2016should}
Marco~Tulio Ribeiro, Sameer Singh, and Carlos Guestrin.
\newblock " why should i trust you?" explaining the predictions of any classifier.
\newblock In \emph{Proceedings of the 22nd ACM SIGKDD international conference on knowledge discovery and data mining}, pages 1135--1144, 2016.

\bibitem[Selvaraju et~al.(2017)Selvaraju, Cogswell, Das, Vedantam, Parikh, and Batra]{selvaraju2017grad}
Ramprasaath~R Selvaraju, Michael Cogswell, Abhishek Das, Ramakrishna Vedantam, Devi Parikh, and Dhruv Batra.
\newblock Grad-cam: Visual explanations from deep networks via gradient-based localization.
\newblock In \emph{Proceedings of the IEEE international conference on computer vision}, pages 618--626, 2017.

\bibitem[Shapley(1953)]{shapley1953value}
Lloyd~S Shapley.
\newblock A value for n-person games.
\newblock \emph{Contributions to the Theory of Games}, 2\penalty0 (28):\penalty0 307--317, 1953.

\bibitem[Shen et~al.(2023)Shen, Cheng, Yang, Li, and Zhang]{shen2023can}
Wen Shen, Lei Cheng, Yuxiao Yang, Mingjie Li, and Quanshi Zhang.
\newblock Can the inference logic of large language models be disentangled into symbolic concepts?
\newblock \emph{arXiv preprint arXiv:2304.01083}, 2023.

\bibitem[Simonyan et~al.(2013)Simonyan, Vedaldi, and Zisserman]{simonyan2013deep}
Karen Simonyan, Andrea Vedaldi, and Andrew Zisserman.
\newblock Deep inside convolutional networks: Visualising image classification models and saliency maps.
\newblock \emph{arXiv preprint arXiv:1312.6034}, 2013.

\bibitem[Sundararajan et~al.(2020)Sundararajan, Dhamdhere, and Agarwal]{sundararajan2020shapley}
Mukund Sundararajan, Kedar Dhamdhere, and Ashish Agarwal.
\newblock The shapley taylor interaction index.
\newblock In \emph{International Conference on Machine Learning}, pages 9259--9268. PMLR, 2020.

\bibitem[Tsai et~al.(2022)Tsai, Yeh, and Ravikumar]{tsai2022faith}
Che-Ping Tsai, Chih-Kuan Yeh, and Pradeep Ravikumar.
\newblock Faith-shap: The faithful shapley interaction index.
\newblock \emph{arXiv preprint arXiv:2203.00870}, 2022.

\bibitem[Wang et~al.(2021{\natexlab{a}})Wang, Lin, Zhang, Zhu, and Zhang]{wangxin2021interpreting}
Xin Wang, Shuyun Lin, Hao Zhang, Yufei Zhu, and Quanshi Zhang.
\newblock Interpreting attributions and interactions of adversarial attacks.
\newblock In \emph{2021 {IEEE/CVF} International Conference on Computer Vision, {ICCV} 2021, Montreal, QC, Canada, October 10-17, 2021}, pages 1075--1084. {IEEE}, 2021{\natexlab{a}}.

\bibitem[Wang et~al.(2021{\natexlab{b}})Wang, Ren, Lin, Zhu, Wang, and Zhang]{wangxin2021unified}
Xin Wang, Jie Ren, Shuyun Lin, Xiangming Zhu, Yisen Wang, and Quanshi Zhang.
\newblock A unified approach to interpreting and boosting adversarial transferability.
\newblock In \emph{9th International Conference on Learning Representations, {ICLR} 2021, Virtual Event, Austria, May 3-7, 2021}, 2021{\natexlab{b}}.

\bibitem[Yosinski et~al.(2015)Yosinski, Clune, Nguyen, Fuchs, and Lipson]{yosinski2015understanding}
Jason Yosinski, Jeff Clune, Anh Nguyen, Thomas Fuchs, and Hod Lipson.
\newblock Understanding neural networks through deep visualization.
\newblock \emph{arXiv preprint arXiv:1506.06579}, 2015.

\bibitem[Zeiler and Fergus(2014)]{zeiler2014visualizing}
Matthew~D Zeiler and Rob Fergus.
\newblock Visualizing and understanding convolutional networks.
\newblock In \emph{European conference on computer vision}, pages 818--833. Springer, 2014.

\bibitem[Zhang et~al.(2021{\natexlab{a}})Zhang, Zhang, Zhou, Bao, Huo, Chen, Cheng, Wu, and Zhang]{zhang2021building}
Die Zhang, Hao Zhang, Huilin Zhou, Xiaoyi Bao, Da~Huo, Ruizhao Chen, Xu~Cheng, Mengyue Wu, and Quanshi Zhang.
\newblock Building interpretable interaction trees for deep {NLP} models.
\newblock In \emph{Thirty-Fifth {AAAI} Conference on Artificial Intelligence, {AAAI} 2021, Thirty-Third Conference on Innovative Applications of Artificial Intelligence, {IAAI} 2021, The Eleventh Symposium on Educational Advances in Artificial Intelligence, {EAAI} 2021, Virtual Event, February 2-9, 2021}, pages 14328--14337. {AAAI} Press, 2021{\natexlab{a}}.

\bibitem[Zhang et~al.(2021{\natexlab{b}})Zhang, Li, Ma, Li, Xie, and Zhang]{zhang2021dropout}
Hao Zhang, Sen Li, YinChao Ma, Mingjie Li, Yichen Xie, and Quanshi Zhang.
\newblock Interpreting and boosting dropout from a game-theoretic view.
\newblock In \emph{International Conference on Learning Representations}, 2021{\natexlab{b}}.

\bibitem[Zhang et~al.(2021{\natexlab{c}})Zhang, Xie, Zheng, Zhang, and Zhang]{zhang2021interpreting}
Hao Zhang, Yichen Xie, Longjie Zheng, Die Zhang, and Quanshi Zhang.
\newblock Interpreting multivariate shapley interactions in dnns.
\newblock In \emph{Proceedings of the AAAI Conference on Artificial Intelligence}, volume~35, pages 10877--10886, 2021{\natexlab{c}}.

\bibitem[Zhang et~al.(2022)Zhang, Wang, Ren, Cheng, Lin, Wang, and Zhu]{zhangquanshi2022proving}
Quanshi Zhang, Xin Wang, Jie Ren, Xu~Cheng, Shuyun Lin, Yisen Wang, and Xiangming Zhu.
\newblock Proving common mechanisms shared by twelve methods of boosting adversarial transferability.
\newblock \emph{arXiv preprint arXiv:2207.11694}, 2022.

\bibitem[Zhou et~al.(2023)Zhou, Zhang, Deng, Liu, Shen, Chan, and Zhang]{zhouhuilin2023generalization}
Huilin Zhou, Hao Zhang, Huiqi Deng, Dongrui Liu, Wen Shen, Shih-Han Chan, and Quanshi Zhang.
\newblock Concept-level explanation for the generalization of a dnn.
\newblock \emph{arXiv preprint arXiv:2302.13091}, 2023.

\end{thebibliography}

\end{document}